\documentclass{article}

\usepackage[preprint]{neurips_2025}

\usepackage[utf8]{inputenc} 
\usepackage[T1]{fontenc}    
\usepackage{hyperref}       
\usepackage{url}            
\usepackage{booktabs}       
\usepackage{amsfonts}       
\usepackage{nicefrac}       
\usepackage{microtype}      
\usepackage{xcolor}         
\usepackage{graphicx}
\usepackage{algorithm}
\usepackage{wrapfig}
\usepackage{algorithmic}
\usepackage{extarrows}
\usepackage{colortbl}
\usepackage{multirow}
\usepackage{pifont}
\usepackage{makecell}
\usepackage{subfig}

\title{ClinKD: Cross-Modal Clinical Knowledge Distiller For Multi-Task Medical Images}

\begin{document}

\author{
Hongyu Ge$^{1}$ $\quad$
Longkun Hao$^{2}$\footnotemark[1] $\quad$
Zihui Xu$^{3}$\footnotemark[1] $\quad$
Zhenxin Lin$^{4}$ $\quad$
Bin Li$^{5}$\footnotemark[2]$\quad$
Shoujun Zhou$^{5}$\footnotemark[2]$\quad$ \\
\textbf{Hongjin Zhao}$^{6}$$\quad$
\textbf{Yihang Liu}$^{3}$ \\
\textsuperscript{\rm 1} The Hong Kong University of Sciences and Technology, Guangzhou $\quad$\\
\textsuperscript{\rm 2} Beihang University
\textsuperscript{\rm 3} Shandong University
\textsuperscript{\rm 4} Hubei University\\
\textsuperscript{\rm 5} Shenzhen Institutes of Advanced Technology, Chinese Academy of Sciences \\
\textsuperscript{\rm 6} Australian National University
}
\maketitle
\renewcommand{\thefootnote}{\fnsymbol{footnote}} 
\footnotetext[1]{Equal contributions in no particular order.} 
\footnotetext[2]{Corresponding Authors; Emails:\{b.li2, sj.zhou\}@siat.ac.cn}

\begin{abstract}
    \label{abstract}
Medical Visual Question Answering (Med-VQA) represents a critical and challenging subtask within the general VQA domain. Despite significant advancements in general VQA, multimodal large language models (MLLMs) still exhibit substantial limitations when handling multi-task VQA scenarios. These limitations manifest through erroneous spatial localization and misinterpretation of medical images, which primarily arise from two fundamental issues: inadequate image-text alignment and insufficient domain-specified knowledge for medical applications. To address these issues, we introduce the Cross-Modal \textbf{Clin}ical \textbf{K}nowledge \textbf{D}istiller (\textbf{ClinKD}), an innovative framework designed to enhance image-text alignment and establish more effective medical knowledge transformation mechanisms, which enables MLLMs to perform better even when lacking prior medical knowledge. Our extensive experimental evaluations demonstrate that the ClinKD achieves state-of-the-art performance on several datasets which are challenging for the Med-VQA task. The results indicate that our approach not only significantly improves image-text alignment but also effectively enables MLLMs to adapt to the medical knowledge. The source code for ClinKD is available at: \url{https://github.com/overloadedHenry/ClinKD}.
\begin{figure}[h]
    \centering
    \includegraphics[scale=0.4]{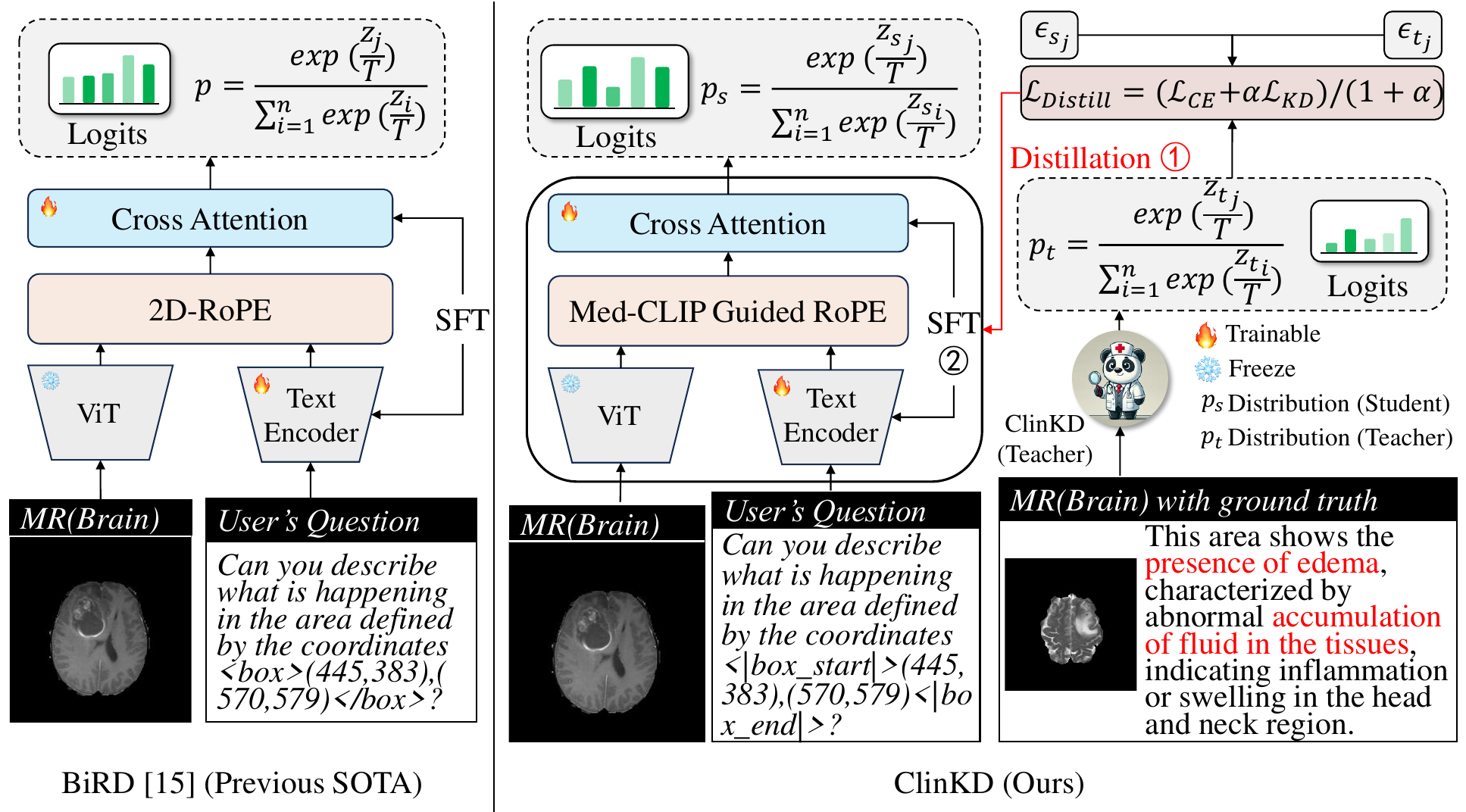}
    \caption{The left part shows that the BiRD~\cite{huang2024BiRD} only utilizes traditional supervised fine-tuning (SFT) and 2D-RoPE~\cite{su2023roformerenhancedtransformerrotary} to obtain the capabilities of understanding and grounding medical images. The right part is our method which uses Med-CLIP Guided RoPE for enhancing image-text alignment and Adaptive Confidence-Margin Curriculum Pseudo-KD for filling the gap of medical knowledge in ClinKD. During the SFT procedure, we will use Reflective Correction Training to enable answers with low semantic similarity or incorrect groundings to be reused.}
    \label{fig:compare}
    \vspace{-2em}
\end{figure}
        
\end{abstract}
\section{Introduction}
\vspace{-1em}
Medical Visual Question Answering (Med-VQA)~\cite{Lin_2023,Abacha2019VQAMedOO, Hasan2018OverviewOI} is a critical domain for applying MLLMs~\cite{yang2024modelmergingllmsmllms, liang2024comprehensivesurveyguidemultimodal, caffagni-etal-2024-revolution} to medical image analysis~\cite{li2024towards}. Recent advancements in MLLMs have enabled them to achieve preliminary capabilities in image analysis, annotation, and user instruction compliance~\cite{chen2025rllavaimprovingmedvqaunderstanding}. Previous studies have made multiple attempts~\cite{yang2025llmmedqaenhancingmedicalquestion,  kumar2024medvisionllamaleveragingpretrainedlarge, he2024parameterefficientfinetuningmedicalmultimodal,liu2024hcllmhistoricalconstrainedlargelanguage,li2024llava, huang2024BiRD} to enhance the performance of general-purpose MLLMs in Med-VQA. 

For instance, LLaVA-Med~\cite{li2024llava} extends instruction-tuning techniques for MLLMs to the medical domain. This work applied instruction tuning and curated image-text datasets to align visual-textual features in LLaVA~\cite{llava}, enabling LLaVA-Med to recognize and analyze medical images. However, this system lacked fine-grained image understanding, causing erroneous spatial localization when questioned about specific pathological locations~\cite{gai2024enhancing, huang2024BiRD}. 

Subsequently, the BiRD~\cite{huang2024BiRD}, a model that incorporated fine-grained image recognition by developing a biomedical refer-and-ground instruction-tuning dataset, addresses the absence of grounding capabilities~\cite{you2023ferretrefergroundgranularity, zhang2024ferretv2improvedbaselinereferring} in LLaVA-Med's work~\cite{li2024llava}. Nevertheless, as shown in Figure~\ref{fig:compare}, due to its reliance on traditional supervised fine-tuning (instruction-tuning)~\cite{sft1, sft2, sft3} and conventional 2D Rotary Position Embedding (RoPE)~\cite{su2023roformerenhancedtransformerrotary}, BiRD tends to generate erroneous spatial localization and misinterpretation of medical images because of incomplete image-text alignment~\cite{alignment1, alignment2}. Moreover, the visual encoder of BiRD is constrained by the lack of medical knowledge, resulting in difficulties in medical knowledge adaptation~\cite{adaptation, lin2025healthgptmedicallargevisionlanguage}.

To overcome these limitations, we propose a medical distillation framework named ClinKD to boost capabilities in the Med-VQA domain. Specifically, our method mainly focuses on two issues: incomplete image-text alignment and the professional medical knowledge gap between the general-purpose MLLMs and medical domain that requires finer-grained medical knowledge~\cite{medical_fine-grained, zeng2024visualorientedfinegrainedknowledgeediting}.

For the first issue, we propose Med-CLIP Guided Rotary Position Embedding (MCG-RoPE), a novel position embedding method that focuses on the inter-modal and intra-modal intervals during joint image-text training, while ensuring the equivalence of context intervals immediately adjacent to images. Compared to traditional RoPE~\cite{su2023roformerenhancedtransformerrotary}, the MCG-RoPE uses the distinct index intervals to better capture cross-modal information, thereby achieving more effective image-text alignment.

For the second issue, we propose \textbf{Clin}ical \textbf{K}nowledge \textbf{D}istiller (\textbf{ClinKD}) which consists of Adaptive Confidence-Margin Curriculum Pseudo-KD (Pseudo-KD) and Reflective Correction Training. Pseudo-KD provides prior medical knowledge by using pseudo-labels~\cite{pseudolabels, chen2023mixedpseudolabelssemisupervised}, filling the gap of prior medical knowledge so that the model can better adapt to medical knowledge during the supervised fine-tuning. Reflective Correction Training allows samples with low semantic similarity to be sent for training again after being enhanced by MLLMs.

Our contributions are summarized as follows:  
\begin{itemize}
    \item We introduce the ClinKD, an innovative distiller designed to establish more effective medical knowledge adaptation mechanisms. The code and checkpoint will be released.
    \item For solving incomplete
image-text alignment, we propose the Med-CLIP Guided Rotary Position Embedding (MCG-RoPE). The MCG-RoPE utilizes the distinct inter-modal and intra-modal features to improve image-text alignment.
    \item For bridging medical knowledge gap
between general-purpose MLLMs and specialized medical
applications, we propose the Adaptive Confidence-Margin Curriculum Pseudo-KD, filling the medical knowledge gap before supervised fine-tuning. 
    \item Extensive experimental results show that the proposed ClinKD achieves competitive or superior performance on test set of diverse tasks and datasets.
\end{itemize}


\begin{figure*}[h!]
\centering
\includegraphics[width=\textwidth]{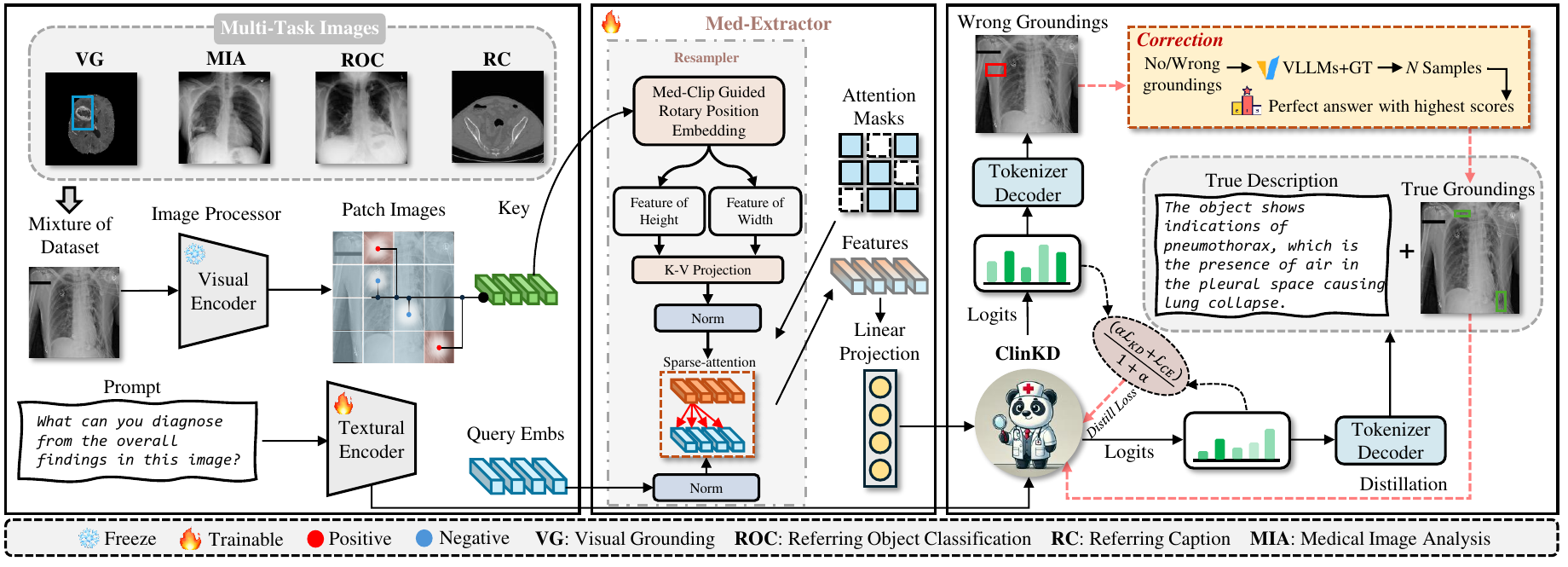} 
\caption{Overview of our ClinKD's framework: The framework begins with multi-task medical images being segmented into patches. The med-extractor module with Med-CLIP Guided Rotary Position Embedding extracts dimensional features from the patched image embeddings. The ClinKD system undergoes distillation before supervised fine-tuning (SFT). During the SFT, answers with low semantic similarity or incorrect groundings are prioritized for retraining. When responding to queries, the framework selects the answer with the highest score, ensuring accurate medical image analysis and interpretation.}
\label{fig:framework}
\vspace{-2em}
\end{figure*}

\section{Related Work}
\vspace{-1em}
\subsection{Biomedical Instruction-Tuning Datasets}
\label{dataset}
\vspace{-1em}
Many Med-VQA datasets such as VQA-RAD~\cite{lau2018dataset} and SLAKE~\cite{liu2021slake} contain only medical images and simple QA pairs. Specifically, these QA pairs do not have the fine-grained descriptions of pathological features. This limitation constrains the ability of MLLMs to perform accurate spatial localization. To boost and evaluate MLLMs' capabilities in the Med-VQA domain, many biomedical instruction-tuning datasets are designed.

For example, the LLaVA-Med-qa0.2k~\cite{zhang2025biomedclipmultimodalbiomedicalfoundation} contains 50 medical images sampled from several modalities such as CT, Angiography and PET and 193 questions according to these medical images. The medical QA pairs are generated by GPT-4 based on the metadata. This dataset is designed to evaluate the capabilities of describing medical images, specifically the biomedical features.

Unlike the LLaVA-Med-qa0.2k~\cite{zhang2025biomedclipmultimodalbiomedicalfoundation}, the Med-GRIT-270k~\cite{huang2024BiRD} includes multi-task fine-grained medical QA pairs especially for grounding spatial locations. It is designed to enhance the model's capabilities in referring and grounding. This multi-task dataset comprises four distinct tasks: Visual Grounding (VG), Referring Object Classification (ROC), Referring Captioning (RC), and Medical Image Analysis (MIA). 
The VG task evaluates the model's capability of accurately matching text descriptions to corresponding image regions. The RC task assesses the model's capability to recognize specific image areas and generate descriptive captions for them. The ROC task examines the model's understanding of textual information related to particular image regions and their associated visual details. The MIA task evaluates the model's comprehension of medical images and their multi-modal context.
All data is sourced from eight modalities: CT, MR, X-ray, PET, Endoscopy, Dermoscopy, Fundus, and Ultrasound. Utilizing data from these eight modalities allows for a more comprehensive evaluation of the model's capabilities.
\vspace{-2em}
\subsection{Multimodal Large Language Models (MLLMs) for Biomedicine}
\vspace{-1em}
With the development of MLLMs, an increasing number of approaches have been applied to the medical field, such as BioMedGPT~\cite{Zhang_2024}, LLaVA-Med~\cite{li2024llava}, and BiRD~\cite{huang2024BiRD}. These methods drive the development of MLLMs in the biomedical domain from different perspectives. For instance, LLaVA-Med~\cite{li2024llava} applied instruction fine-tuning to the medical imaging field and built a medical question-answering system based on LLaVA~\cite{llava}, which provided reasonable responses by processing user-submitted images and instructions. However, LLaVA-Med did not support fine-grained interactions with medical images, meaning that the system cannot precisely locate pathological regions within the images. To address the lack of referring and grounding capabilities, the BiRD~\cite{huang2024BiRD}, which is based on Qwen-VL~\cite{bai2023qwenvlversatilevisionlanguagemodel}, leveraged a finer-grained medical dataset. However, the ViT of BiRD was not trained by sufficient medical data before being frozen, resulting in a gap in medical knowledge domain. Compared with their work~\cite{huang2024BiRD, li2024llava}, our method modifies the position embedding method and introduces an effective Confidence-Margin Curriculum Pseudo-KD framework.
\vspace{-1em}
\subsection{Rotary Position Embedding (RoPE)}
\vspace{-1em}
The introduced RoPE~\cite{su2023roformerenhancedtransformerrotary} makes the relative position features of the context in the text more easily captured by LLMs. Further, the RoPE-Mixed~\cite{heo2024rotarypositionembeddingvision} introduced mixed learnable frequencies to RoPE, enabling it to handle diagonal directions and making RoPE itself learnable. It showed more efficiency when dealing with diagonal features of the image. The VideoRoPE~\cite{wei2025videoropemakesgoodvideo} extended the diagonal layout and variable frequency to the video domain, demonstrating its advantages in video analysis tasks. Unlike existing methods, our approach focuses on 2D multi-task medical images where capturing semantic information is challenging. We utilize cross-modal intervals to enable the model to better distinguish modal information, thereby improving image-text alignment.

\vspace{-1em}
\section{Methodology}
\label{method}
\begin{figure*}[h]
\centering
\includegraphics[width=0.9\textwidth]{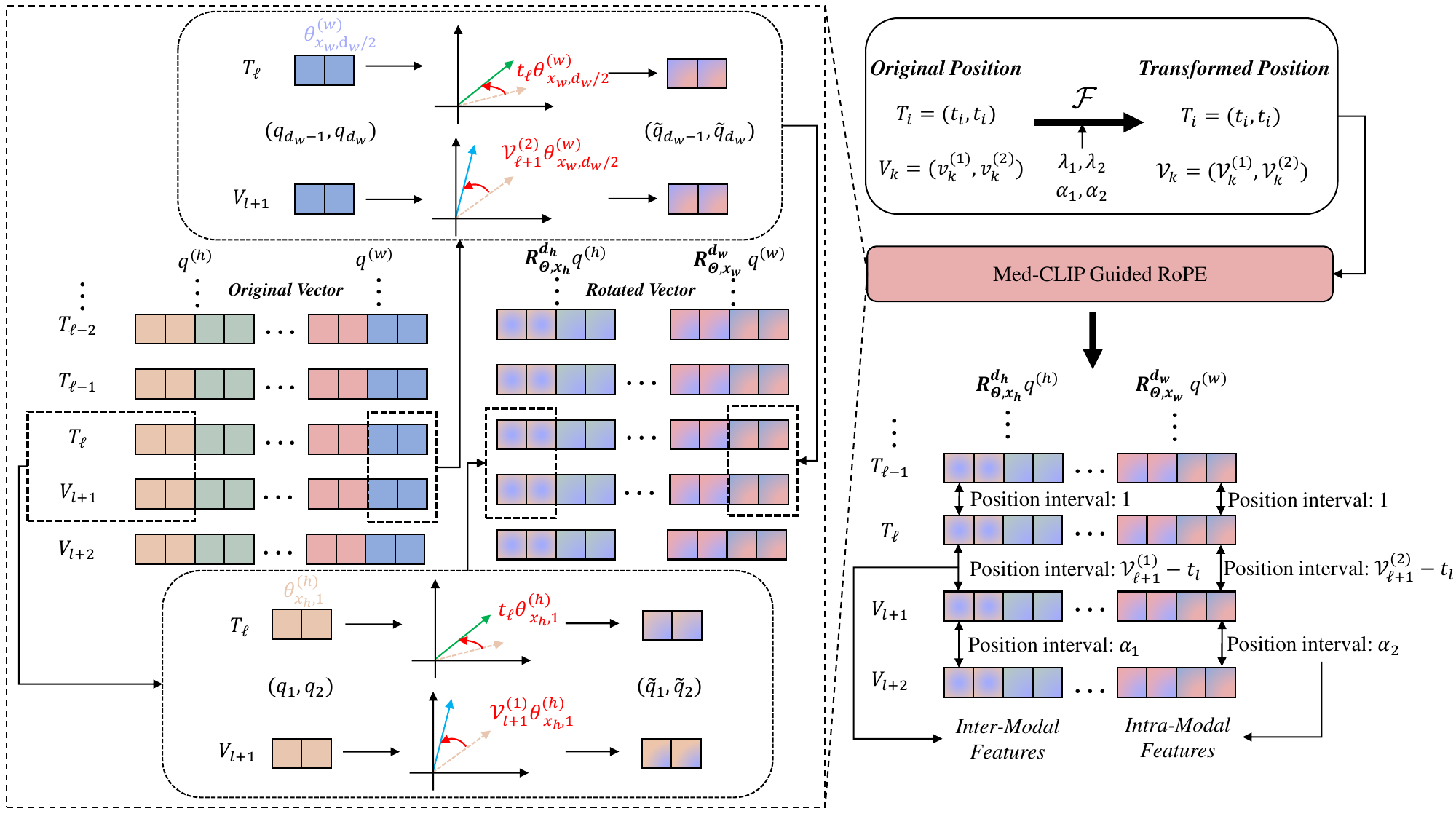} 

\caption{The visualized example of Med-CLIP Guided Rotary Position Embedding. The original vector will be rotated by different angles according to the transformed position indexes. The inter-modal and intra-modal intervals are different so that the Med-CLIP Guided Rotary Position Embedding can better align image feature and text feature.}
\label{fig:3}
\vspace{-1em}
\end{figure*}
As shown in Figure~\ref{fig:framework}, we aim at enhancing the ClinKD's performance by mainly leveraging Med-CLIP Guided Rotary Position Embedding, Pseudo-KD, Reflective Correction Training and Semantic-Aware Selective Generation. The proposed Med-CLIP Guided Rotary Position uses the distinct inter-modal and intra-modal space to improve image-text alignment. The distiller provides prior medical knowledge to enhance the model's capability of medical knowledge adaptation. In the inference procedure, the Semantic-Aware Selective Generation will allow the model to choose the answer with the best score.
\vspace{-1em}
\subsection{Med-CLIP Guided Rotary Position Embedding}
\vspace{-1em}
The standard 2D Rotary Position Embedding (RoPE)~\cite{su2023roformerenhancedtransformerrotary} does not explicitly account for cross-modal relationships between image and text tokens. To address this, we introduce the Med-CLIP Guided Rotary Position Embedding (MCG-RoPE), which deal with tokens from different modals by considering splited queries to improve image-text alignment. Figure~\ref{fig:3} illustrates an example of our approach.
\paragraph{Distinct Inter-Modal Intervals.}  
Consider a sequence of position indices for a text–image–text pattern:
\begin{equation}
S_{\mathrm{idx}}
=
\bigl(T_1, T_2, \dots, T_\ell,\;
V_{\ell+1}, \dots, V_{\ell+wh},\;
T_{\ell+wh+1}, \dots, T_{\ell+wh+k}\bigr),
\end{equation}
where each text index \(T_i=(t_i,t_i)\in\mathbb{R}^2\) and each image index \(V_j=(v^{(1)}_j, v^{(2)}_j)\in\mathbb{R}^2\).  
We apply an affine transform \(\mathcal{F}\) to each image index so that intra-modal and inter-modal gaps become equal along both axes:
\begin{equation}
\mathcal{F}\bigl(v_i^{(1)},v_i^{(2)}\bigr)
=
\bigl(\alpha_1\,v_i^{(1)} + \lambda_1,\;\alpha_2\,v_i^{(2)} + \lambda_2\bigr)
=:\mathcal{V}_i,
\end{equation}
with scaling factors \(\alpha_1,\alpha_2>0\) and offsets \(\lambda_1,\lambda_2\in\mathbb{R}\).  
By enforcing a pronounced interval difference between text and image tokens, the model is guided to focus on fine-grained visual details in medical images while maintaining clear separation from surrounding text.  




\paragraph{Distinct Interval-Based Rotation.}  
Let \(q\in\mathbb{R}^{d_h+d_w}\) be a query/key vector, split as
\begin{equation}
q = \bigl[q^{(h)}; q^{(w)}\bigr],
\quad
q^{(h)}\in\mathbb{R}^{d_h},\;
q^{(w)}\in\mathbb{R}^{d_w}.
\end{equation}
By processing the height and width dimensions separately, the model can capture both coarse inter-modal alignment and fine intra-modal structure.  

Define frequency scales for each half of the embedding:
\begin{equation}
\omega_{i}^{(h)} = 10\,000^{-\frac{2i}{d_h}},
\quad
\omega_{i}^{(w)} = 10\,000^{-\frac{2i}{d_w}},
\quad
i = 1,\dots,\tfrac{d_h}{2}\;\text{or}\;\tfrac{d_w}{2}.
\end{equation}

Given the transformed positions \(x_h\) and \(x_w\), the rotation angles satisfy
\begin{equation}
\theta_{x_h,i}^{(h)} = x_h\,\omega_i^{(h)},
\quad
\theta_{x_w,i}^{(w)} = x_w\,\omega_i^{(w)}.
\end{equation}

We construct block-diagonal rotation matrices
\begin{equation}
\label{rotation}
\mathbf{R}_{\Theta,x_h}^{d_h}
=\mathrm{diag}\bigl(R_{\theta_{x_h,1}^{(h)}},\dots,
R_{\theta_{x_h,\frac{d_h}{2}}^{(h)}}\bigr),
\quad
\mathbf{R}_{\Theta,x_w}^{d_w}
=\mathrm{diag}\bigl(R_{\theta_{x_w,1}^{(w)}},\dots,
R_{\theta_{x_w,\frac{d_w}{2}}^{(w)}}\bigr),
\end{equation}
where each \(2\times2\) block is
\begin{equation}
R_{\theta}
=
\begin{pmatrix}
\cos\theta & -\sin\theta\\
\sin\theta & \cos\theta
\end{pmatrix}.
\end{equation}

Finally, the transformed vector is
\begin{equation}
\label{eq:mcgrope}
\widetilde{q}
=
\bigl[
\mathbf{R}_{\Theta,x_h}^{d_h}\,q^{(h)};
\,
\mathbf{R}_{\Theta,x_w}^{d_w}\,q^{(w)}
\bigr].
\end{equation}

By emphasizing distinct interval gaps and splitting the query into separate height and width components, MCG-RoPE enables the model to more effectively capture fine-grained features in medical images and achieve precise image–text alignment.
The difference between 2D-RoPE and MCG-RoPE is shown in Appendix.
\vspace{-1em}
\subsection{Adaptive Confidence-Margin Curriculum Pseudo-KD}
\vspace{-1em}
In order to mitigate the negative impact of noisy pseudo-labels and more effectively guide the student model, we propose an \emph{Adaptive Confidence–Margin Curriculum Pseudo-KD} strategy.  This approach extends standard pseudo-label distillation by (1) weighting each sample’s distillation loss according to both the teacher’s confidence and the margin between top logits, and (2) employing a simple curriculum learning schedule that gradually relaxes the confidence threshold over training.
\vspace{-1em}
\paragraph{Confidence and Margin Computation.}  
Given an input example \(x\), the teacher network produces logits \(z_t = f_t(x)\).  We compute the softmax probabilities
\begin{equation}
p_{t_i}(x) \;=\; \frac{\exp\bigl(z_{t_i}\bigr)}{\sum_j \exp\bigl(z_{t_j}\bigr)} \,.
\end{equation}
We define the \emph{confidence} and \emph{margin} measures as
\begin{equation}
C(x) \;=\; \max_i p_{t_i}(x)
\quad,\quad
\Delta(x) \;=\; p_{t_{(1)}}(x) \;-\; p_{t_{(2)}}(x)\,,
\end{equation}
where \(p_{(1)}\) and \(p_{(2)}\) denote the largest and second-largest probabilities, respectively.

\paragraph{Curriculum Threshold Schedule.}  
Let \(t\in[0,1]\) denote the normalized training progress (i.e., fraction of total steps completed).  We set an initial confidence threshold \(\tau_0\) and a minimum threshold \(\tau_{\min}\).  The threshold decays linearly:
\begin{equation}
\tau(t) \;=\; \tau_{\min} \;+\; (\tau_0 - \tau_{\min}) \,\bigl(1 - t\bigr)\,.
\end{equation}

\paragraph{Adaptive Weighting Function.}  
We combine confidence and margin into a per-sample weight:
\begin{equation}
w(x) \;=\;
\begin{cases}
\bigl[C(x)\bigr]^{\gamma}\,\bigl[\Delta(x)\bigr]^{\beta}, 
& C(x) \;\ge\; \tau(t)\,,\\
0, & \text{otherwise},
\end{cases}
\end{equation}
where \(\gamma,\beta\in(0,1]\) control sensitivity to confidence and margin.

\paragraph{Weighted Distillation Loss.}  
With student logits \(z_s\) and temperature \(T\), the weighted KL-divergence distillation loss is
\begin{equation}
\mathcal{L}_{\mathrm{KD}}
=\frac{1}{N}\sum_{i=1}^N w(x_i)\;
\mathbb{D}_{KL}\Bigl(softmax\bigl(z_{t_i}/T\bigr)\,\big\|\,softmax\bigl(z_{s_i}/T\bigr)\Bigr)\,.
\end{equation}

\paragraph{Overall Objective.}  
We combine the standard cross-entropy loss \(L_{\mathrm{CE}}\) with our adaptive pseudo-KD term:
\begin{equation}
\label{distillation}
\mathcal{L}_{Distill} \;=\; \dfrac{\mathcal{L}_{\mathrm{CE}}
\;+\;\alpha\,\mathcal{L}_{\mathrm{KD}}}{1+\alpha}
\quad,\quad
\alpha\in[0,1]\,.
\end{equation}

\subsection{Reflective Correction Training}
\label{gpt4o}
\vspace{-1em}
In this part, during each round of supervised fine‐tuning, we record the model’s prediction  
\(\hat{\mathbf{y}} = (\mathbf{t}_o, \mathbf{c})\)  
and the corresponding ground‐truth label  
\(\mathbf{y} = (\mathbf{t}_{gt}, \mathbf{c})\).  
We then compute a weighted cosine similarity to assess semantic alignment and, if necessary, invoke GPT-4o to “correct” the output in order to enrich label diversity. The procedure is as follows:

First, map the text labels into a \(d\)-dimensional embedding space to obtain $\mathbf{e}_o,\mathbf{e}_{gt} \in \mathbb{R}^d$, while retaining the shared anchor‐box coordinate vector  
\(\mathbf{c} \in \mathbb{R}^4\).

Next, define the weighted cosine similarity as  
\begin{equation}
S(\mathbf{e}_o,\mathbf{e}_{gt};\mathbf{c})
=
\frac{
\mu\,\langle \mathbf{e}_o, \mathbf{e}_{gt}\rangle
+
\nu\,\langle \mathbf{c},\mathbf{c}\rangle
}{
\sqrt{\bigl(\mu\,\|\mathbf{e}_o\|^2 + \nu\,\|\mathbf{c}\|^2\bigr)\,
      \bigl(\mu\,\|\mathbf{e}_{gt}\|^2 + \nu\,\|\mathbf{c}\|^2\bigr)}
}\,,
\label{eq:weighted-cosine}
\end{equation}
where \(\mu,\nu>0\) weight the contribution of semantic similarity and coordinate consistency, respectively; \(\langle\cdot,\cdot\rangle\) denotes the inner product and \(\|\cdot\|\) the Euclidean norm. We typically choose \(\mu>\nu\) to emphasize textual alignment. Let the similarity threshold be \(\tau = 0.8\).

We then compare \(S\) against \(\tau\) and decide whether to trigger GPT-4o correction:
\begin{equation}
\hat{\mathbf{y}}' =
\begin{cases}
\mathrm{GPT4o\_Correct}\bigl(\hat{\mathbf{y}},\,\mathbf{y}\bigr), 
& S(\mathbf{e}_o,\mathbf{e}_{gt};\mathbf{c}) < \tau,\\[6pt]
\hat{\mathbf{y}}, 
& S(\mathbf{e}_o,\mathbf{e}_{gt};\mathbf{c}) \ge \tau,
\end{cases}
\label{eq:correction}
\end{equation}
where \(\mathrm{GPT4o\_Correct}(\cdot)\) leaves the anchor‐box \(\mathbf{c}\) unchanged and revises only the text to match the ground‐truth intent.
\vspace{-1em}
\subsection{Semantic-Aware Selective Generation}
\vspace{-1em}
The responses generated by large language models exhibit randomness, and relying on a single inference may result in missing the optimal answer.
Semantic-Aware Selective Generation means that instead of directly using the model's single-round generation for evaluation, we allow the model to generate multiple samples. We then utilize the CLIP models to score the alignment between the generated text and image. Finally, we reorder the outputs based on these scores and select the one with the highest score as the final result. The pseudo code is shown in Algorithm.

\vspace{-1em}
\section{Experiments}
\label{exper}
\vspace{-1em}
\subsection{Datasets}
\vspace{-1em}
We train ClinKD on the \textbf{Med-GRIT-270k~\cite{huang2024BiRD}} dataset, and evaluate the performance on Med-GRIT-30k and \textbf{LLaVA-Med-qa0.2k~\cite{zhang2025biomedclipmultimodalbiomedicalfoundation}} dataset.

We alse test the few-shot performance on \textbf{VQA-RAD~\cite{lau2018dataset}}, \textbf{SLAKE}~\cite{liu2021slake} and \textbf{Path-VQA}~\cite{he2020pathvqa}. For more information, please refer to Appendix.
\vspace{-1em}
\subsection{Evaluation Metrics}
\vspace{-1em}
We use Recall@0.5 for Visual Grounding (VG), Recall for Referring Object Classification (ROC), SPICE~\cite{anderson2016spicesemanticpropositionalimage} for Referring Caption (RC), and mBMR~\cite{huang2024BiRD} for Medical Image Analysis (MIA). The final overall result is computed as the mean of the evaluation scores across these four tasks.

For the evaluation on LLaVA-Med-qa0.2k~\cite{li2024llava} which has 0.2k QA pairs of MIA task, we use mBMR as the metric.

In few-shot experiments, we use BLEU-1 and F1-score to evaluate models' performance. 

\vspace{-1em}
\subsection{Analysis}
\subsubsection{Comparison with the State-of-the-Art Methods}
\textbf{Evaluation on the Med-GRIT-Test30k~\cite{ye2023samed2d20m, cheng2023sammed2d}.} As shown in Table~\ref{tab:new_sota_data}, we compare ClinKD with BiRD~\cite{huang2024BiRD} and LLaVA-Med~\cite{li2024llava}. Results show that ClinKD achieves scores of 67.51\%, 82.35\%, 70.56\%, and 65.69\% in the VG, ROC, RC, and MIA tasks respectively, outperforming other comparisons by a clear margin in all tasks. On average, ClinKD outperforms BiRD~\cite{huang2024BiRD} and LLaVA-Med~\cite{li2024llava} by 14.87\% and 66.00\% respectively. ClinKD's advantages lie in its more adequate image-text alignment and the prior medical knowledge provided by Pseudo-KD.\\
\textbf{Evaluation on the LLaVA-Med-qa0.2k~\cite{zhang2025biomedclipmultimodalbiomedicalfoundation}.} Table~\ref{tab:new_sota_data} represents the models' performance on LLaVA-Med-qa0.2k~\cite{zhang2025biomedclipmultimodalbiomedicalfoundation}. Our ClinKD model demonstrates superior performance on this dataset, outperforming both LLaVA-Med (by 3.50\%) and BiRD (by 2.48\%). The relatively low mBMR scores can be attributed to the notable language style discrepancies observed between the training and testing datasets.
\begin{table*}[h]
\centering
\vspace{-1em}
\resizebox{\textwidth}{!}{%
\begin{tabular}{c|c|cccc|c}
\toprule
\textbf{Model} & \textbf{Test dataset} & \textbf{VG (Recall@0.5)} & \textbf{ROC (Recall)} & \textbf{RC (SPICE)} & \textbf{MIA (mBMR)} & \textbf{Average} \\ \midrule
LLaVA-Med~\cite{li2024llava}  & Med-GRIT-Test30k~\cite{ye2023samed2d20m, cheng2023sammed2d} & 0 & 2.75 & 8.18 & 11.20 & 5.53 \\ 

BiRD~\cite{huang2024BiRD} & Med-GRIT-Test30k~\cite{ye2023samed2d20m, cheng2023sammed2d} & 53.92 & 65.33 & 55.23 & 52.17 & 56.66 \\ 
\rowcolor[RGB]{255,245,235}
\textbf{ClinKD (Ours)} & Med-GRIT-Test30k~\cite{ye2023samed2d20m, cheng2023sammed2d} & \textbf{67.51} & \textbf{82.35} & \textbf{70.56} & \textbf{65.69} & \textbf{71.53} \\
\midrule
LLaVA-Med~\cite{li2024llava}  & LLaVA-Med-qa0.2k~\cite{zhang2025biomedclipmultimodalbiomedicalfoundation} & - & - & - & 20.04 & - \\
BiRD~\cite{huang2024BiRD} & LLaVA-Med-qa0.2k~\cite{zhang2025biomedclipmultimodalbiomedicalfoundation} & - & - & - & 21.06 & - \\ 

ClinKD (Ours)  & LLaVA-Med-qa0.2k~\cite{zhang2025biomedclipmultimodalbiomedicalfoundation} & - & - & - & \textbf{23.54} & - \\
\bottomrule
\end{tabular}%
}
\caption{Comparison with LLaVA-Med~\cite{li2024llava} and BiRD (previous SOTA)~\cite{huang2024BiRD}. The best scores are shown in bold.}
\vspace{-2em}
\label{tab:new_sota_data}
\end{table*}
\begin{table*}[h]
\centering
\resizebox{\textwidth}{!}{%
\begin{tabular}{c|ccc|cccc|c}
\toprule
\multirow{2}{*}{\textbf{Model}} & \multicolumn{3}{c|}{\textbf{Proposed Methods}}  & \multirow{2}{*}{\textbf{VG (Recall@0.5)}} & \multirow{2}{*}{\textbf{ROC (Recall)}} & \multirow{2}{*}{\textbf{RC (SPICE)}} & \multirow{2}{*}{\textbf{MIA (mBMR)}} & \multirow{2}{*}{\textbf{Average}} \\ 
\cline{2-4}
&MCG-RoPE & Pseudo-KD &SASG & & & & & \\ 
\midrule
Qwen2-VL~\cite{wang2024qwen2vlenhancingvisionlanguagemodels}  &\ding{55}&\ding{55}&\ding{55} & 55.43 \textcolor{blue}{(+1.51)} & 68.33 \textcolor{blue}{(+3.00)} & 57.39 \textcolor{blue}{(+2.16)} & 62.23 \textcolor{blue}{(+10.06)} & 61.37 \textcolor{blue}{(+4.71)} \\
\midrule
\multirow{4}{*}{BiRD~\cite{huang2024BiRD}}  &\ding{55} &\ding{55}&\ding{55} & 53.92 & 65.33 & 55.23 & 52.17 & 56.66 \\ 
~  &\ding{51}&\ding{55}&\ding{55}& 64.03 \textcolor{blue}{(+10.11)} & 72.74 \textcolor{blue}{(+7.41)} & 61.32 \textcolor{blue}{(+6.09)} & 59.63 \textcolor{blue}{(+7.46)} & 64.44 \textcolor{blue}{(+7.78)} \\
~ &\ding{55}&\ding{51}&\ding{55}& 66.33 \textcolor{blue}{(+12.41)} & 79.52  \textcolor{blue}{(+14.19)} & 67.21 \textcolor{blue}{(+11.98)} & 64.32 \textcolor{blue}{(+12.15)} & 69.35 \textcolor{blue}{(+12.69)} \\
~ &\ding{55}&\ding{55}&\ding{51}& 55.89 \textcolor{blue}{(+1.97)} & 67.84 \textcolor{blue}{(+2.51)} & 54.32 \textcolor{blue}{(-0.91)} & 51.27 \textcolor{blue}{(-0.90)} & 57.33 \textcolor{blue}{(+0.67)} \\
\midrule
\rowcolor[RGB]{255,245,235}
\textbf{ClinKD (Ours)} &\ding{51}&\ding{51}&\ding{51} & \textbf{67.51} \textcolor{blue}{(+13.59)} & \textbf{82.35} \textcolor{blue}{(+17.02)} & \textbf{70.56} \textcolor{blue}{(+15.33)} & \textbf{65.69} \textcolor{blue}{(+13.52)} & \textbf{71.53} \textcolor{blue}{(+14.87)} \\
\bottomrule
\end{tabular}%
}
\caption{Ablation study for different models and proposed methods. The value which is blue shows how much the metric scores improved. The best scores are shown in bold. The MCG-RoPE, Pseudo-KD and SASG stand for Med-CLIP Guided RoPE, Adaptive Confidence-Margin Curriculum Pseudo-KD and Semantic-Aware Selective Generation respectively.}
\vspace{-2em}
\label{tab:ablation_data}
\end{table*}
\subsubsection{Ablation study}
\textbf{Effect of Qwen2-VL~\cite{wang2024qwen2vlenhancingvisionlanguagemodels}.} Table \ref{tab:ablation_data} presents the ablation study of  proposed methods. To investigate the impact of model upgrade on the experiment, we applied the same training strategy to Qwen2-VL as we did to BiRD~\cite{huang2024BiRD}. The results show that the model upgrade only has a significant effect on the MIA task, which shows Qwen2-VL~\cite{wang2024qwen2vlenhancingvisionlanguagemodels} outperforms BiRD by 10.06\%.\\
\textbf{Effect of Med-CLIP Guided Rotary Position Embedding.} As shown in Table~\ref{tab:ablation_data}, with adding MCG-RoPE to BiRD, all metrics exhibit significant enhancements, with the average rising from 56.66\% to 64.44\%. This indicates the importance of adequate image-text alignment.\\
\textbf{Effect of Adaptive Confidence-Margin Curriculum Pseudo-KD.} Table~\ref{tab:ablation_data} represents improvements achieved by using Pseudo-KD that leads to enhancements by 12.41\%, 14.19\%, 11.98\% and 12.15\% in the VG, ROC, RC and MIA tasks, respectively. The results illustrate that providing prior medical knowledge by using Pseudo-KD can fill MLLMs' gap of medical knowledge, boosting capabilities of adapting to new medical knowledge.
Figure
\ shows the effect of different values of $\alpha$ in Eq.~(\ref{distillation}). In this experiment, the average performance achieves the best when we set $\alpha=0.5$. Both excessively high and low alpha values can lead to performance degradation. This might be because a low $\alpha$ value prevents ClinKD from learning sufficient prior medical knowledge, while an overly high alpha value tends to make ClinKD generate irrational pseudo-labels that contradict common medical knowledge.\\
\textbf{Effect of Semantic-Aware Selective Generation.} From the results in Table~\ref{tab:ablation_data}, we can see a light enhancement on VG and ROC task. However, there is a slight decrease in the metric scores on the RC and MIA tasks. These results may be caused by the semantic similarity from CLIP models that are affected by complex sentences.

\begin{table*}[h]
\centering

\resizebox{\textwidth}{!}{%
\begin{tabular}{c|c|c|cccccccc|c}
\toprule
\textbf{Task} & \textbf{Metric} &\textbf{Method} & \textbf{CT} & \textbf{MR} & \textbf{X-ray} & \textbf{PET} & \textbf{Endoscopy} & \textbf{Dermoscopy} & \textbf{Fundus} & \textbf{Ultrasound} & \textbf{Average} \\ \midrule

~   & ~ & BiRD~\cite{huang2024BiRD} & 44.47\textcolor{red}{$\pm$0.14} & 29.26\textcolor{red}{$\pm$0.11} & 41.73\textcolor{red}{$\pm$0.16} & 56.46\textcolor{red}{$\pm$0.15} & 53.60\textcolor{red}{$\pm$0.12} & 75.63\textcolor{red}{$\pm$0.17} & 84.15\textcolor{red}{$\pm$0.13} & 46.04\textcolor{red}{$\pm$0.18} & 53.92\textcolor{red}{$\pm$0.14} \\
~   & ~ & BiRD~\cite{huang2024BiRD} + RoPE-Mixed~\cite{heo2024rotarypositionembeddingvision} & 46.90\textcolor{red}{$\pm$0.16} & 53.11\textcolor{red}{$\pm$0.13} & 42.03\textcolor{red}{$\pm$0.17} & 60.84\textcolor{red}{$\pm$0.14} & 53.49\textcolor{red}{$\pm$0.15} & 78.25\textcolor{red}{$\pm$0.11} & 85.16\textcolor{red}{$\pm$0.12} & 52.83\textcolor{red}{$\pm$0.13} & 59.08\textcolor{red}{$\pm$0.19}   \\
\rowcolor[RGB]{237,238,254}
\cellcolor{white}~   & \cellcolor{white}~ & BiRD~\cite{huang2024BiRD} + MCG-RoPE (Ours) & 51.22\textcolor{red}{$\pm$0.14} & 49.40\textcolor{red}{$\pm$0.18} & 45.92\textcolor{red}{$\pm$0.17} & 68.60\textcolor{red}{$\pm$0.13} & 60.37\textcolor{red}{$\pm$0.19} & 88.28\textcolor{red}{$\pm$0.12} & 86.59\textcolor{red}{$\pm$0.15} & 61.88\textcolor{red}{$\pm$0.16} & 64.03\textcolor{red}{$\pm$0.15}   \\
~   & ~ & BiRD~\cite{huang2024BiRD} + $V_{k}D$ ~\cite{Miles_2024_CVPR}  & 58.77\textcolor{red}{$\pm$0.12} & 50.95\textcolor{red}{$\pm$0.16} & 45.69\textcolor{red}{$\pm$0.11} & 65.89\textcolor{red}{$\pm$0.18} & 57.63\textcolor{red}{$\pm$0.14} & 93.32\textcolor{red}{$\pm$0.15} & 79.27\textcolor{red}{$\pm$0.12} & 59.90\textcolor{red}{$\pm$0.15} & 63.93\textcolor{red}{$\pm$0.19}   \\
\rowcolor[RGB]{237,238,254}
\cellcolor{white}~   &\cellcolor{white} ~ & BiRD~\cite{huang2024BiRD} + Pseudo-KD (Ours) & 62.59\textcolor{red}{$\pm$0.15} & 52.86\textcolor{red}{$\pm$0.17} & 47.19\textcolor{red}{$\pm$0.13} & 66.28\textcolor{red}{$\pm$0.16} & 59.84\textcolor{red}{$\pm$0.11} & 93.68\textcolor{red}{$\pm$0.18} & 85.37\textcolor{red}{$\pm$0.14} & 62.87\textcolor{red}{$\pm$0.19} & 66.33\textcolor{red}{$\pm$0.13}   \\
\rowcolor[RGB]{255,245,235}
\cellcolor{white}\multirow{-6}*{\textbf{VG}}   &\cellcolor{white} \multirow{-6}*{Recall@0.5} & ClinKD (Ours) & \textbf{65.52}\textcolor{red}{$\pm$0.13} & \textbf{52.23}\textcolor{red}{$\pm$0.30} & \textbf{48.56}\textcolor{red}{$\pm$0.19} & \textbf{69.25}\textcolor{red}{$\pm$0.12} & \textbf{60.37}\textcolor{red}{$\pm$0.16} & \textbf{94.22}\textcolor{red}{$\pm$0.10} & \textbf{86.59}\textcolor{red}{$\pm$0.14} & \textbf{64.47}\textcolor{red}{$\pm$0.15} & \textbf{67.51}\textcolor{red}{$\pm$0.18} \\ \midrule

~   & ~ &BiRD~\cite{huang2024BiRD}& 34.76\textcolor{red}{$\pm$0.11} & 61.79\textcolor{red}{$\pm$0.15} & 53.74\textcolor{red}{$\pm$0.14} &  -     & 60.40\textcolor{red}{$\pm$0.17} & 96.61\textcolor{red}{$\pm$0.18}     & -     & 84.65\textcolor{red}{$\pm$0.12}     &  65.33\textcolor{red}{$\pm$0.14}  \\
~   & ~ &BiRD~\cite{huang2024BiRD} + RoPE-Mixed~\cite{heo2024rotarypositionembeddingvision} & 39.78\textcolor{red}{$\pm$0.13} & 62.13\textcolor{red}{$\pm$0.19} & 74.27\textcolor{red}{$\pm$0.11} &  -     & 61.33\textcolor{red}{$\pm$0.13} & 96.61\textcolor{red}{$\pm$0.17}     & -     & 84.65\textcolor{red}{$\pm$0.15}     &  69.80\textcolor{red}{$\pm$0.12}  \\
\rowcolor[RGB]{237,238,254}
\cellcolor{white}~   &\cellcolor{white} ~ &BiRD~\cite{huang2024BiRD} + MCG-RoPE (Ours)& 44.47\textcolor{red}{$\pm$0.15} & 64.20\textcolor{red}{$\pm$0.18} & 83.75\textcolor{red}{$\pm$0.11} &  -     & 62.19\textcolor{red}{$\pm$0.14} & 96.61\textcolor{red}{$\pm$0.19}     & -     & 85.22\textcolor{red}{$\pm$0.16}     &  72.74\textcolor{red}{$\pm$0.13}  \\
~   & ~ &BiRD~\cite{huang2024BiRD} + $V_{k}D$~\cite{Miles_2024_CVPR}& 53.19\textcolor{red}{$\pm$0.17} & 69.10\textcolor{red}{$\pm$0.14} & 90.84\textcolor{red}{$\pm$0.19} &  -     & 61.71\textcolor{red}{$\pm$0.10} & \textbf{100}\textcolor{red}{$\pm$0.00}     & -     & 81.15\textcolor{red}{$\pm$0.12}     &  76.00\textcolor{red}{$\pm$0.15}  \\
\rowcolor[RGB]{237,238,254}
\cellcolor{white}~   & \cellcolor{white}~ &BiRD~\cite{huang2024BiRD} + Pseudo-KD (Ours) & 58.47\textcolor{red}{$\pm$0.16} & 69.92\textcolor{red}{$\pm$0.12} & 95.83\textcolor{red}{$\pm$0.18} &  -     & 66.32\textcolor{red}{$\pm$0.11} & \textbf{100}\textcolor{red}{$\pm$0.00}     & -     & 86.58\textcolor{red}{$\pm$0.13}     &  79.52\textcolor{red}{$\pm$0.14}  \\
\rowcolor[RGB]{255,245,235}
\cellcolor{white}\multirow{-6}{*}{\textbf{ROC}}   & \cellcolor{white}\multirow{-6}{*}{Recall} &ClinKD (Ours)& \textbf{61.42}\textcolor{red}{$\pm$0.17} & \textbf{71.30}\textcolor{red}{$\pm$0.19} & \textbf{97.00}\textcolor{red}{$\pm$0.18} & - & \textbf{68.84}\textcolor{red}{$\pm$0.12} &  \textbf{100}\textcolor{red}{$\pm$0.00}   & - & \textbf{95.54}\textcolor{red}{$\pm$0.11} & \textbf{82.35}\textcolor{red}{$\pm$0.15}  \\ \midrule
~   & ~ &BiRD~\cite{huang2024BiRD}& 41.88\textcolor{red}{$\pm$0.12} & 51.69\textcolor{red}{$\pm$0.14} & 37.39\textcolor{red}{$\pm$0.18} & 47.95\textcolor{red}{$\pm$0.15} & 54.07\textcolor{red}{$\pm$0.10} & 77.44\textcolor{red}{$\pm$0.16} & 48.73\textcolor{red}{$\pm$0.13} & 82.65\textcolor{red}{$\pm$0.19} & 55.23\textcolor{red}{$\pm$0.14}    \\
~   & ~ &BiRD~\cite{huang2024BiRD} + RoPE-Mixed~\cite{heo2024rotarypositionembeddingvision}& 43.51\textcolor{red}{$\pm$0.15} & 51.88\textcolor{red}{$\pm$0.18} & 54.61\textcolor{red}{$\pm$0.13} & 52.83\textcolor{red}{$\pm$0.19} & 54.07\textcolor{red}{$\pm$0.11} & 68.15\textcolor{red}{$\pm$0.17} & 53.99\textcolor{red}{$\pm$0.12} & 65.28\textcolor{red}{$\pm$0.14} & 55.54\textcolor{red}{$\pm$0.13}    \\
\rowcolor[RGB]{237,238,254}
\cellcolor{white}~   & \cellcolor{white}~ &BiRD~\cite{huang2024BiRD} + MCG-RoPE (Ours)& 50.73\textcolor{red}{$\pm$0.14} & 53.11\textcolor{red}{$\pm$0.16} & 60.33\textcolor{red}{$\pm$0.12} & 70.16\textcolor{red}{$\pm$0.18} & 54.07\textcolor{red}{$\pm$0.11} & 70.18\textcolor{red}{$\pm$0.19} & 61.52\textcolor{red}{$\pm$0.15} & 70.46\textcolor{red}{$\pm$0.17} & 61.32\textcolor{red}{$\pm$0.13}    \\
~   & ~ &BiRD~\cite{huang2024BiRD} + $V_{k}D$ ~\cite{Miles_2024_CVPR} & 51.33\textcolor{red}{$\pm$0.13} & 54.39\textcolor{red}{$\pm$0.17} & 62.83\textcolor{red}{$\pm$0.19} & 74.31\textcolor{red}{$\pm$0.14} & 54.07\textcolor{red}{$\pm$0.15} & 76.23\textcolor{red}{$\pm$0.12} & 67.89\textcolor{red}{$\pm$0.18} & 72.15\textcolor{red}{$\pm$0.11} & 64.15\textcolor{red}{$\pm$0.16}    \\
\rowcolor[RGB]{237,238,254}
\cellcolor{white}~   &\cellcolor{white} ~ &BiRD~\cite{huang2024BiRD} + Pseudo-KD (Ours)& 52.78\textcolor{red}{$\pm$0.12} & 58.41\textcolor{red}{$\pm$0.15} & 70.39\textcolor{red}{$\pm$0.18} & 75.85\textcolor{red}{$\pm$0.13} & \textbf{55.22}\textcolor{red}{$\pm$0.16} & 81.34\textcolor{red}{$\pm$0.11} & 71.54\textcolor{red}{$\pm$0.17} & 72.15\textcolor{red}{$\pm$0.14} & 67.21\textcolor{red}{$\pm$0.19}    \\
\rowcolor[RGB]{255,245,235}
\cellcolor{white}\multirow{-6}{*}{\textbf{RC}}   &\cellcolor{white} \multirow{-6}{*}{SPICE~\cite{anderson2016spicesemanticpropositionalimage}} &ClinKD (Ours)& \textbf{53.94}\textcolor{red}{$\pm$0.14} & \textbf{58.41}\textcolor{red}{$\pm$0.12} & \textbf{71.47}\textcolor{red}{$\pm$0.18} & \textbf{79.91}\textcolor{red}{$\pm$0.15}& 52.10\textcolor{red}{$\pm$0.11} & \textbf{83.92}\textcolor{red}{$\pm$0.19} & \textbf{84.51}\textcolor{red}{$\pm$0.13} & \textbf{80.19}\textcolor{red}{$\pm$0.16} & \textbf{70.56}\textcolor{red}{$\pm$0.17} \\ \midrule

~  & ~ &BiRD~\cite{huang2024BiRD}& 47.01\textcolor{red}{$\pm$0.14} & 49.35\textcolor{red}{$\pm$0.18} & 37.17\textcolor{red}{$\pm$0.15} & 57.15\textcolor{red}{$\pm$0.12} & 39.91\textcolor{red}{$\pm$0.17} & 72.13\textcolor{red}{$\pm$0.13} & 48.87\textcolor{red}{$\pm$0.16} & 65.78\textcolor{red}{$\pm$0.19} & 52.17\textcolor{red}{$\pm$0.11} \\ 
~  & ~ &BiRD~\cite{huang2024BiRD} + RoPE-Mixed~\cite{heo2024rotarypositionembeddingvision} & 57.10\textcolor{red}{$\pm$0.17} & 50.89\textcolor{red}{$\pm$0.13} & 42.55\textcolor{red}{$\pm$0.16} & 69.28\textcolor{red}{$\pm$0.11} & 41.63\textcolor{red}{$\pm$0.18} & 73.32\textcolor{red}{$\pm$0.14} & 59.63\textcolor{red}{$\pm$0.15} & 61.80\textcolor{red}{$\pm$0.10} & 57.03\textcolor{red}{$\pm$0.19} \\ 
\rowcolor[RGB]{237,238,254}
\cellcolor{white}~  &\cellcolor{white} ~ &BiRD~\cite{huang2024BiRD} + MCG-RoPE (Ours)& 60.53\textcolor{red}{$\pm$0.16} & 53.77\textcolor{red}{$\pm$0.11} & 54.10\textcolor{red}{$\pm$0.18} & 62.01\textcolor{red}{$\pm$0.12} & 43.88\textcolor{red}{$\pm$0.15} & 75.33\textcolor{red}{$\pm$0.14} & 62.89\textcolor{red}{$\pm$0.19} & 64.53\textcolor{red}{$\pm$0.13} & 59.63\textcolor{red}{$\pm$0.17} \\ 
~  & ~ &BiRD~\cite{huang2024BiRD} + $V_{k}D$ ~\cite{Miles_2024_CVPR} & 62.33\textcolor{red}{$\pm$0.10} & 53.77\textcolor{red}{$\pm$0.15} & 52.27\textcolor{red}{$\pm$0.19} & 62.01\textcolor{red}{$\pm$0.16} & 47.29\textcolor{red}{$\pm$0.13} & 75.33\textcolor{red}{$\pm$0.17} & 68.04\textcolor{red}{$\pm$0.12} & 70.31\textcolor{red}{$\pm$0.18} & 61.42\textcolor{red}{$\pm$0.14} \\ 
\rowcolor[RGB]{237,238,254}
\cellcolor{white}~  & \cellcolor{white}~ &BiRD~\cite{huang2024BiRD} + Pseudo-KD (Ours)& 66.85\textcolor{red}{$\pm$0.11} & 55.73\textcolor{red}{$\pm$0.14} & 52.27\textcolor{red}{$\pm$0.17} & 64.74\textcolor{red}{$\pm$0.15} & \textbf{50.58}\textcolor{red}{$\pm$0.12} & 78.21\textcolor{red}{$\pm$0.19} & 70.02\textcolor{red}{$\pm$0.13} & \textbf{76.16}\textcolor{red}{$\pm$0.18} & 64.32\textcolor{red}{$\pm$0.16} \\ 
\rowcolor[RGB]{255,245,235}
\cellcolor{white}\multirow{-6}{*}{\textbf{MIA}}  &\cellcolor{white} \multirow{-6}{*}{mBMR~\cite{huang2024BiRD}} &ClinKD (Ours)& \textbf{66.85}\textcolor{red}{$\pm$0.13} & \textbf{56.82}\textcolor{red}{$\pm$0.16} & \textbf{54.01}\textcolor{red}{$\pm$0.18} & \textbf{67.08}\textcolor{red}{$\pm$0.14} & \textbf{50.58}\textcolor{red}{$\pm$0.11} & \textbf{81.00}\textcolor{red}{$\pm$0.15} & \textbf{70.02} \textcolor{red}{$\pm$0.17} & 76.05\textcolor{red}{$\pm$0.19} & \textbf{65.69}\textcolor{red}{$\pm$0.12} \\ \midrule

~ & ~ &BiRD~\cite{huang2024BiRD}& 43.03\textcolor{red}{$\pm$0.12} & 48.02\textcolor{red}{$\pm$0.11} & 42.51\textcolor{red}{$\pm$0.17} & 53.85\textcolor{red}{$\pm$0.19} & 51.99\textcolor{red}{$\pm$0.14} & 80.45\textcolor{red}{$\pm$0.16} & 60.58\textcolor{red}{$\pm$0.13} & 69.78\textcolor{red}{$\pm$0.15} &  -     \\ 
~ & ~ &BiRD~\cite{huang2024BiRD} + RoPE-Mixed~\cite{heo2024rotarypositionembeddingvision}& 46.82\textcolor{red}{$\pm$0.17} & 54.50\textcolor{red}{$\pm$0.12} & 53.37\textcolor{red}{$\pm$0.15} & 60.98\textcolor{red}{$\pm$0.13} & 52.63\textcolor{red}{$\pm$0.18} & 79.08\textcolor{red}{$\pm$0.10} & 66.26\textcolor{red}{$\pm$0.19} & 66.14\textcolor{red}{$\pm$0.11} &  -     \\ 
\rowcolor[RGB]{237,238,254}
\cellcolor{white}~ & \cellcolor{white}~ &BiRD~\cite{huang2024BiRD} + MCG-RoPE (Ours)& 51.74\textcolor{red}{$\pm$0.13} & 55.12\textcolor{red}{$\pm$0.19} & 61.03\textcolor{red}{$\pm$0.12} & 66.92\textcolor{red}{$\pm$0.17} & 55.13\textcolor{red}{$\pm$0.14} & 82.60\textcolor{red}{$\pm$0.15} & 70.33\textcolor{red}{$\pm$0.11} & 70.52\textcolor{red}{$\pm$0.18} &  -     \\ 
~ & ~ &BiRD~\cite{huang2024BiRD} + $V_{k}D$~\cite{Miles_2024_CVPR} & 56.41\textcolor{red}{$\pm$0.16} & 57.05\textcolor{red}{$\pm$0.11} & 62.91\textcolor{red}{$\pm$0.19} & 67.40\textcolor{red}{$\pm$0.14} & 55.18\textcolor{red}{$\pm$0.18} & 86.22\textcolor{red}{$\pm$0.10} & 71.73\textcolor{red}{$\pm$0.15} & 70.88\textcolor{red}{$\pm$0.13} &  -     \\ 
\rowcolor[RGB]{237,238,254}
\cellcolor{white}~ &\cellcolor{white} ~ &BiRD~\cite{huang2024BiRD} + Pseudo-KD (Ours)& 60.17\textcolor{red}{$\pm$0.14} & 59.23\textcolor{red}{$\pm$0.16} & 66.42\textcolor{red}{$\pm$0.12} & 68.96\textcolor{red}{$\pm$0.19} & \textbf{57.99}\textcolor{red}{$\pm$0.13} & 88.31\textcolor{red}{$\pm$0.15} & 75.64\textcolor{red}{$\pm$0.17} & 74.44\textcolor{red}{$\pm$0.10} &  -     \\ 
\rowcolor[RGB]{255,245,235}
\cellcolor{white}\multirow{-6}{*}{\textbf{Average}} & \cellcolor{white}\multirow{-6}{*}{-} &ClinKD (Ours)& \textbf{61.93}\textcolor{red}{$\pm$0.15} & \textbf{59.69}\textcolor{red}{$\pm$0.18} & \textbf{67.76}\textcolor{red}{$\pm$0.19} & \textbf{72.08}\textcolor{red}{$\pm$0.13} & 57.97\textcolor{red}{$\pm$0.16}  & \textbf{89.79}\textcolor{red}{$\pm$0.11} & \textbf{80.37}\textcolor{red}{$\pm$0.14} & \textbf{79.06}\textcolor{red}{$\pm$0.12} & -  \\ \bottomrule
\end{tabular}%
}
\caption{Comparison with other SOTA methods on Med-GRIT-Test30k~\cite{ye2023samed2d20m, cheng2023sammed2d}. We evaluate the performance on eight modalities.}
\label{tab:detail_res}
\vspace{-1em}
\end{table*}
\vspace{-1em}
\subsubsection{Comparison with Other Methods}
\vspace{-1em}
\textbf{Comparison with BiRD + RoPE-Mixed~\cite{heo2024rotarypositionembeddingvision}.}
The RoPE-Mixed uses frequencies for both axes as learnable network parameters, effectively handling image features in diagonal direction. It achieves the best performance in different visual scenarios~\cite{ade20k, zhou2018semanticunderstandingscenesade20k, coco}. Hence we compare our MCG-RoPE with RoPE-Mixed. As shown in Table~\ref{tab:detail_res}, by adding RoPE-Mixed to BiRD~\cite{huang2024BiRD}, great achievements take place on all modalities (CT, MR, X-ray, etc.). Compared with RoPE-Mixed~\cite{heo2024rotarypositionembeddingvision}, our MCG-RoPE outperforms it by 4.95\%, 2.94\%, 5.78\% and 2.60\% on VG, ROC, RC and MIA tasks, respectively. For more detailed performance on the 8 modalities, MCG-RoPE still achieves the best improvements most of the time.\\
\textbf{Comparison with BiRD + $V_{k}D$~\cite{Miles_2024_CVPR}.}
$V_{k}D$ is a novel method that can improve knowledge distillation. This work utilizes a projection layer for maximizing the knowledge transfer to the student backbone. Many experiments~\cite{DLGAN_CVPR, spkd, pkd, label-kd} have been made to prove the best performance of this method. Compared with BiRD with $V_kD$ in the Table~\ref{tab:detail_res}, Pseudo-KD shows its advantages on multi-task fine-grained Med-VQA scenarios, outperforming $V_kD$ almost on all modalities. 
\vspace{-1em}
\subsubsection{Few-shot Experiments}
Table \ref{tab:few-shot} shows the results of our additional experiments for testing few-shot performance. The evaluation scores shows that our ClinKD achieves the best performance on all datasets. It is can be seen that the BiRD which also utilizes Rotary Position Embedding outperforms LLaVA-Med~\cite{li2024llava} and Med-Flamingo~\cite{med-fla}. This may be caused by the special meaning of 2D-RoPE. For our ClinKD, the modified Med-CLIP Guided RoPE enhances the position information between image tokens and text tokens, enabling better image-text alignment.

\begin{table*}[h]
\centering
\resizebox{0.9\textwidth}{!}{%
\begin{tabular}{c|c|cccc}
\toprule
\textbf{Dataset} & \textbf{Metric} & \textbf{LLaVA-Med~\cite{li2024llava}} & \textbf{BiRD~\cite{huang2024BiRD}} & \textbf{Med-Flamingo~\cite{med-fla}} &\textbf{\cellcolor[RGB]{255,245,235}ClinKD}\\
\midrule
VQA-RAD~\cite{lau2018dataset} &F1 &42.91 &46.12 & 41.79 &\cellcolor[RGB]{255,245,235}\textbf{59.21}\\
\midrule
~ & BLEU-1& 43.35 &48.23 & 43.27&\cellcolor[RGB]{255,245,235}\textbf{57.09}\\
\multirow{-2}{*}{SLAKE~\cite{liu2021slake}} &F1 &44.28 &49.25 & 45.86 &\cellcolor[RGB]{255,245,235}\textbf{58.64}\\
\midrule
~ & BLEU-1& 43.35 &48.23 & 41.98&\cellcolor[RGB]{255,245,235}\textbf{58.63}\\
\multirow{-2}{*}{Path-VQA~\cite{he2020pathvqa}} &F1 &44.28 &49.25  & 42.56&\cellcolor[RGB]{255,245,235}\textbf{59.21}\\
\bottomrule
\end{tabular}
}
\caption{Comparison with LLaVA-Med~\cite{li2024llava} and BiRD (previous SOTA)~\cite{huang2024BiRD}. The best scores are shown in bold.}
\label{tab:few-shot}
\end{table*}
\vspace{-1em}
\subsubsection{Case Study}
\begin{figure}[htbp]
    \centering
    \subfloat[Case study on Med-GRIT-Test30~\cite{ye2023samed2d20m, cheng2023sammed2d}.]{
        \label{fig:visual1}\includegraphics[width=0.5\textwidth]{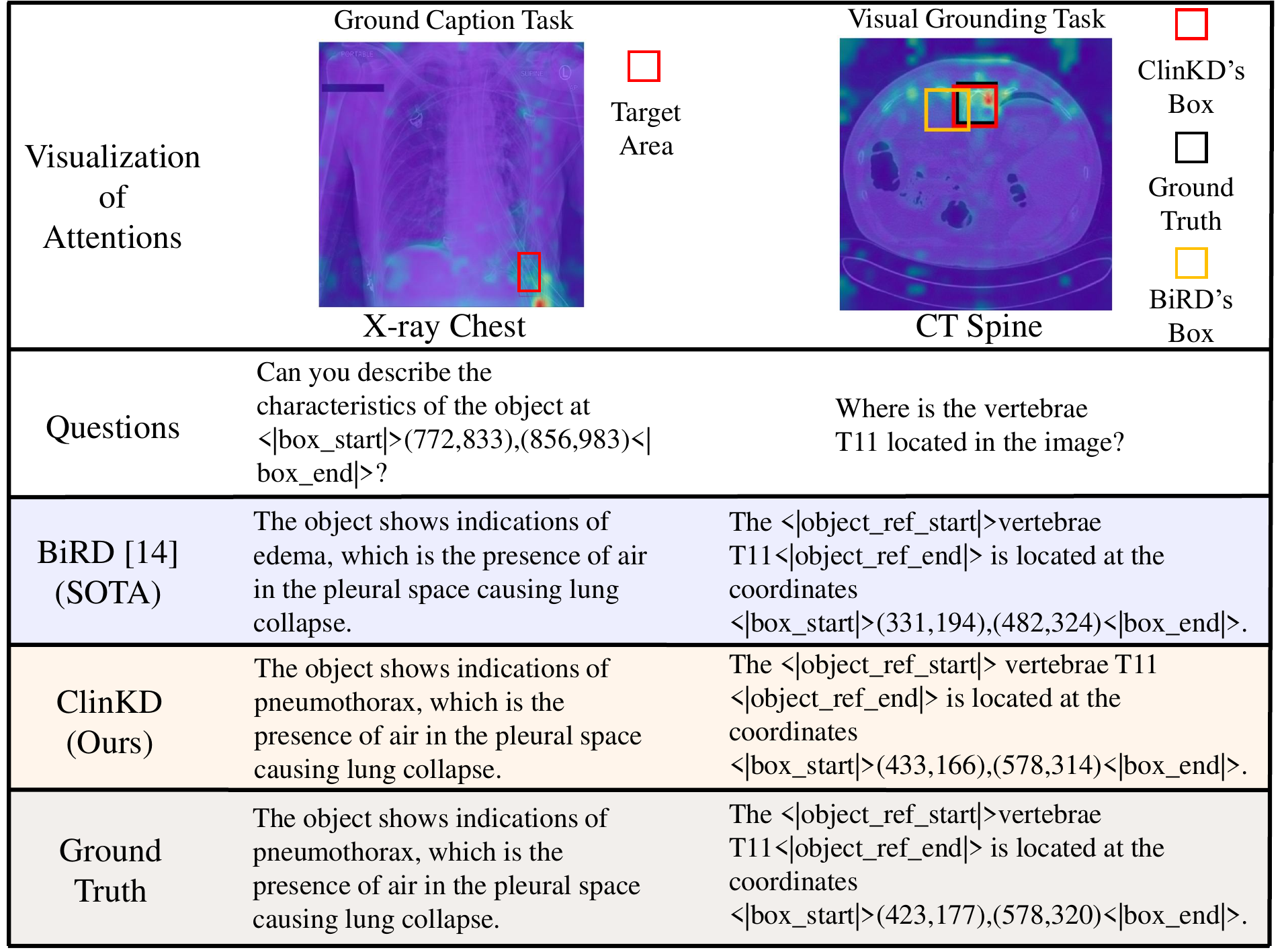}
    }
    \subfloat[Case study on VQA-RAD~\cite{lau2018dataset}.]{
        \label{fig:rad}\includegraphics[width=0.5\textwidth]{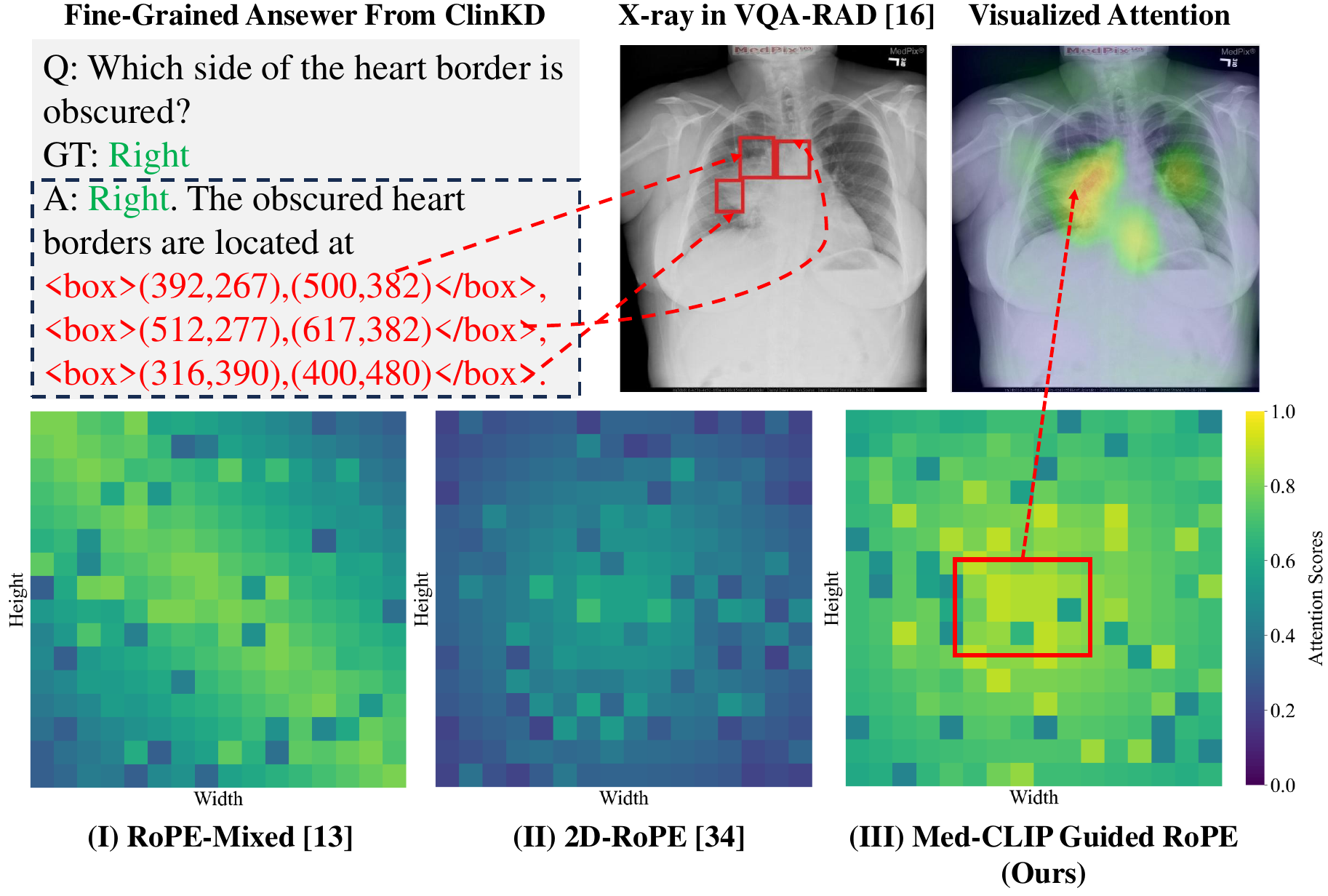}
    }
    \caption{Case Studies on Med-GRIT-Test30~\cite{ye2023samed2d20m, cheng2023sammed2d} and VQA-RAD~\cite{lau2018dataset}.}
    \label{fig:overall}
\end{figure}

\textbf{Case study on Med-GRIT-Test30~\cite{ye2023samed2d20m, cheng2023sammed2d}.}
Figure~\ref{fig:visual1} illustrates the visualization of the attention map and a comparison among BiRD~\cite{huang2024BiRD}, ClinKD (ours), and the ground truth. The grounding of ClinKD is more accurate because MCG-RoPE can improve image-text alignment, and at the same time, Pseudo-KD enables ClinKD to better adapt to medical knowledge.\\
\textbf{Case study on VQA-RAD.}
As shown in Figure~\ref{fig:rad}, we conducted an experiment on VQA-RAD~\cite{lau2018dataset}. Despite the relatively simple annotations, ClinKD provides fine-grained answers with three accurate groundings, demonstrating its precision in response.

To assess the models' attention to pathological features, we conducted a comparative analysis of three RoPE variants. The results indicate that 2D-RoPE~\cite{su2023roformerenhancedtransformerrotary} tends to show low attention on pathological regions. In contrast, RoPE-Mixed~\cite{heo2024rotarypositionembeddingvision} demonstrates a greater emphasis on diagonal features. Notably, our Med-CLIP Guided RoPE specifically targets and enhances the model's attention to pathological regions, thereby facilitating more accurate alignment between text and images, especially in the identification of relevant medical features.
\vspace{-1em}
\section{Conclusion}
In this work, we propose ClinKD, a novel medical distiller centered on Adaptive Confidence-Margin Curriculum Pseudo-KD and Reflective Correction Training. In addition, we propose the Med-CLIP Guided RoPE to improve image-text alignment. They are designed to enhance the performance on fine-grained multi-task datasets in Med-VQA domain. Extensive experiments demonstrate that our Med-CLIP Guided RoPE achieves superior image-text alignment, while the Adaptive Confidence-Margin Curriculum Pseudo-KD effectively bridges prior medical knowledge gaps. These mechanisms synergistically enhance the model's medical knowledge adaptation capabilities. 

In future work, further refinements to the distillation process, along with the incorporation of more comprehensive medical knowledge and higher-quality datasets, could significantly improve model's robustness and reduce the model's reliance on manually labeled data.


{
    \small
    \bibliographystyle{plain}
    \bibliography{neurips_2025}
}

\end{document}